\documentclass[pdflatex,sn-mathphys-num]{sn-jnl}

\usepackage[T1]{fontenc}%
\usepackage{graphicx}%
\usepackage{multirow}%
\usepackage{amsmath,amssymb,amsfonts}%
\usepackage{amsthm}%
\usepackage[title]{appendix}%
\usepackage{xcolor}%
\usepackage{textcomp}%
\usepackage{manyfoot}%
\usepackage{booktabs}%
\usepackage{array}%
\usepackage{url}%
\usepackage{algorithm}%
\usepackage{algorithmicx}%
\usepackage{algpseudocode}%
\usepackage{listings}%

\theoremstyle{thmstyleone}%
\theoremstyle{thmstyletwo}%

\theoremstyle{thmstylethree}%

\newcommand{\best}[1]{\textbf{#1}}
\newcommand{\xseq}{X_{\mathrm{tar}}}
\newcommand{\xcov}{X_{\mathrm{cov}}}

\newcommand{\loss}{\mathcal{L}}

\begin{document}

\title[Plug-in Gate for Transformer Forecasters]{A Lightweight Plug-in Gate for Transformer-Based Time-Series Forecasters}

\author[1]{\fnm{Hongkai} \sur{Zhuang}}

\author*[1]{\fnm{Tao} \sur{Huang}}\email{huang-tao@mju.edu.cn}

\author[1]{\fnm{Chen} \sur{Hou}}

\affil*[1]{\orgdiv{School of Computer and Big Data}, \orgname{Minjiang University}, \orgaddress{\street{No. 200 Xiyuangong Road}, \city{Fuzhou}, \postcode{350108}, \state{Fujian}, \country{China}}}

\abstract{
Covariate-rich time-series forecasting requires deciding how external variables enter the target forecasting path. Existing Transformer-based forecasters usually build a covariate representation and pass it to the encoder without an explicit admission stage. This paper studies pre-encoder covariate admission as an input-side interface that regulates that representation immediately before encoder processing. We implement the interface with a lightweight representation-level pre-encoder gate that assigns sigmoid scores to representation units, and we also study a usage-regularized variant that penalizes average admission. The interface is evaluated as a plug-in module for TimeXer, Inverted Transformer (iTransformer), and Patch Time Series Transformer (PatchTST) under a zero-extra-tuning protocol, where each gated model inherits the corresponding baseline configuration. Experiments on the \textit{Electricity Transformer Temperature minute-level} (\textit{ETTm1} and \textit{ETTm2}) datasets, \textit{Traffic}, \textit{Energy}, and \textit{influenza-like illness} (\textit{ILI}) include paired forecasting comparisons, gate-placement ablation, initialization ablation, controlled covariate-admission analysis, and a variance inflation factor (VIF)-informed permutation feature importance (PFI) diagnostic case study. In the tested settings, the gate is competitive with the corresponding baselines, and the usage penalty reduces average admission scores while keeping forecasting errors close to the unpenalized TimeXer setting.
}


\maketitle

\section{Introduction}

Long-term time-series forecasting is central to energy management, transportation planning, industrial monitoring, and public-health analysis. In these settings, the future value of a target variable is shaped by both the target history and external covariates. Load indicators, neighboring sensors, periodic markers, and public-health signals can each contribute information that is not recoverable from the target series alone. The modeling question is therefore not only how to model the target history, but also how to regulate the entry of covariate information into the forecasting path.

Covariates are not uniformly useful. Some channels contribute independent information; others repeat information already present in the target history or in other channels. In multivariate benchmarks, this redundancy is common because variables are drawn from coupled sensors, correlated physical processes, or shared reporting systems. The issue studied here is whether a Transformer-based forecaster should expose an explicit admission step at the boundary where the covariate representation enters the encoder.

Pre-encoder admission is different from removing variables before training or explaining a trained model after prediction. Offline feature selection changes the input set, while post-hoc analysis only describes a trained model. The design question here is architectural: whether a Transformer-based forecasting model can include a lightweight input-side interface that learns how strongly its covariate representation should be admitted before encoder processing.

Recent long-term forecasting models have improved sequence representation through efficient attention, decomposition, frequency-domain modeling, simple linear baselines, and temporal-variation modeling \cite{zhou2021informer,wu2021autoformer,zhou2022fedformer,zeng2023transformers,wu2022timesnet}. PatchTST segments each channel into temporal patches and preserves channel independence \cite{nie2022time}. iTransformer inverts the conventional tokenization scheme and represents each variable history as one token \cite{liu2024itransformer}. TimeXer focuses on forecasting with covariates by using separate target and covariate embeddings with cross-attention \cite{wang2024timexer}. These designs improve temporal modeling, variable representation, or covariate interaction, but they still leave admission of the resulting covariate representation implicit at the encoder boundary. The narrower interface question is whether a shared representation-level gate can regulate the covariate representation immediately before encoder processing.

We instantiate the interface as a lightweight representation-level pre-encoder gate that can be attached to Transformer-based backbones. After a baseline constructs its covariate representation and before that representation enters the encoder, the gate computes a score for each representation unit, converts the scores into weights, and reweights the representation through element-wise multiplication. The reweighted representation then follows the original encoder and forecasting pipeline. The same two-layer multilayer perceptron (MLP) scoring rule is used at this interface for each evaluated Transformer-based backbone, while the encoder and prediction head remain inherited from the baseline. We also study a usage-regularized variant that penalizes average admission under a controlled usage proxy. This extension tests whether the same interface can reduce admission while keeping forecasting errors close to the unpenalized setting.

The empirical study is organized around five questions. First, under inherited baseline configurations, does a pre-encoder admission gate preserve or improve paired forecasting accuracy? Second, does the pre-encoder placement matter compared with a gate applied after encoder processing or fixed attenuation? Third, how sensitive is the gate to the initial admission probability? Fourth, can an explicit usage penalty control average admission? Fifth, in a representative redundant-covariate setting, how do learned gate weights relate to VIF and PFI diagnostics? These questions keep the scope aligned with the proposed interface: the paper evaluates a pre-encoder covariate-admission mechanism rather than a new forecasting backbone.

The main contributions are as follows.
\begin{itemize}
    \item We formulate pre-encoder covariate admission as a plug-in interface that can be attached to Transformer-based backbones, where a baseline's covariate representation is regulated before encoder processing rather than removed by offline feature selection.
    \item We introduce a lightweight pre-encoder gate and a usage-regularized variant that uses a controlled usage penalty to reduce average covariate admission without redesigning the forecasting backbone.
    \item We evaluate the interface through paired forecasting comparisons, placement-and-learnability ablation, initialization ablation, controlled covariate-admission analysis, and a VIF-informed PFI diagnostic case study.
\end{itemize}

\section{Related Work}

\subsection{Forecasting Backbones}

Long-term forecasting backbones can be grouped by how they represent temporal and variable information. Earlier neural models such as Long- and Short-Term Time-series Network (LSTNet) combine convolutional and recurrent components to capture local and long-term temporal patterns in multivariate series \cite{lai2018modeling}. Transformer-based models then introduced several backbone designs for long-horizon forecasting: Informer reduces the cost of self-attention for long sequences \cite{zhou2021informer}, Autoformer combines decomposition with an auto-correlation mechanism \cite{wu2021autoformer}, Frequency Enhanced Decomposed Transformer (FEDformer) moves part of the modeling into the frequency domain \cite{zhou2022fedformer}, and TimesNet represents temporal variation through two-dimensional transformations \cite{wu2022timesnet}. At the same time, Decomposition-Linear (DLinear) showed that simple linear models remain strong baselines for long-term forecasting \cite{zeng2023transformers}.

Another group of models changes how variables are represented. Crossformer explicitly models cross-dimension dependency in multivariate forecasting \cite{zhang2023crossformer}. PatchTST represents a series as patch tokens and keeps channel independence, so each variable is modeled as a univariate sequence with shared weights \cite{nie2022time}. iTransformer represents each variable history as one token and applies attention over variables \cite{liu2024itransformer}. These methods mainly address temporal representation, variable representation, or cross-variable dependency inside the backbone. The present paper focuses on a different design choice: regulating the covariate representation before the first encoder layer.

\subsection{Covariate Fusion and Reweighting}

A second line of work studies how the target variable should interact with external or auxiliary variables. Dual-stage attention-based recurrent neural network (DA-RNN) uses input attention to select relevant driving series and temporal attention to select relevant hidden states for prediction \cite{qin2017dual}. Temporal Fusion Transformer (TFT) uses variable selection networks and gating components to handle heterogeneous covariates in multi-horizon forecasting \cite{lim2021temporal}. TimeXer is the closest recent Transformer representative for long-term forecasting with covariates. It separates target and covariate embeddings and uses cross-attention to transfer covariate information to the target representation \cite{wang2024timexer}. More generally, attention mechanisms can reweight tokens inside a network \cite{vaswani2017attention}, and stochastic gate methods have also been used for feature selection \cite{yamada2020feature}.

However, attention, variable selection, and feature selection address different placements in the forecasting pipeline. DA-RNN and TFT model feature relevance within recurrent or multi-horizon forecasting architectures. Stochastic gate feature selection targets feature selection before model prediction. TimeXer models covariate interaction through cross-attention after constructing its covariate representation. These mechanisms show that covariate relevance can be modeled, while the placement studied here is a shared admission point immediately before the encoder receives the baseline representation. The present paper focuses on a learnable admission score at that pre-encoder boundary, and the proposed gate is a lightweight implementation of that input-side interface.

\subsection{Covariate Usage and Redundancy Diagnostics}

Covariate admission is also related to cost-sensitive or budgeted feature use, where a model balances predictive utility against the number of variables admitted into a decision process. In this paper, we use a controlled usage penalty as a benchmark-level proxy for covariate usage. This setting is useful for testing whether the proposed interface can reduce average covariate admission while preserving forecasting accuracy. It also separates the interface question from offline feature selection: instead of selecting a fixed subset before training, the model learns continuous admission scores jointly with the forecasting objective.

Covariates in real-world forecasting datasets often arise from coupled sensors or related physical quantities. If one covariate can be largely explained by the others, it adds limited independent information and can increase redundant representation learning. VIF is a standard statistic for diagnosing linear redundancy among explanatory variables \cite{o2007caution}. In this paper, it is used to indicate whether the covariate space contains channels that are easy to reconstruct from other channels.

VIF is a redundancy diagnostic for the covariate space, whereas the forecasting question is how a neural model should regulate covariate information during prediction. This motivates an interface-level comparison between statistical redundancy diagnostics and neural covariate usage. Existing forecasting backbones mainly improve temporal or variable representation, while covariate-aware models mainly improve how external variables are represented and interacted inside the backbone. This paper studies an explicit, backbone-preserving interface that regulates the covariate representation before the first encoder layer. The representation-level pre-encoder gate instantiated in this paper evaluates that interface with the same scoring mechanism across the tested backbones.

\section{Proposed Method}

The proposed method keeps the forecasting backbone unchanged and focuses on the pre-encoder regulation interface. We instantiate this interface with a small learnable gate placed at the pre-encoder boundary after the baseline constructs its covariate representation. The target branch, encoder, prediction objective, horizon setting, and downstream forecasting head are inherited from the corresponding backbone. The gate is therefore a minimal trainable mechanism for testing whether explicit pre-encoder covariate admission is useful.

\subsection{Problem Formulation}

Let $\xseq \in \mathbb{R}^{B \times L \times 1}$ denote the historical target sequence for a batch of size $B$ and look-back length $L$. Let $\xcov \in \mathbb{R}^{B \times L \times C}$ denote the $C$ raw covariates. The goal is to forecast $Y \in \mathbb{R}^{B \times H \times 1}$ for horizon $H$.

We treat the target variable and covariates as distinct inputs. The baseline first constructs its covariate representation. At the pre-encoder boundary, let $E_g$ denote the representation supplied to the gate and let $N_c$ denote the number of representation units in $E_g$. Thus, $C$ counts the raw covariate channels in the input space, whereas $N_c$ counts the representation units exposed to the gate. The two quantities need not be equal, because the gate operates on the representation-level interface rather than on raw covariate identities. This keeps the formulation generic across Transformer-based backbones with different covariate layouts. The gate is applied to $E_g$ before it is passed to the unchanged encoder.

This formulation keeps the forecasting task unchanged. The target history, covariate history, and prediction horizon are the same as in the corresponding baseline setting. The added operation is the learned scaling of the covariate representation before it enters the encoder. A low gate score therefore means reduced admission of a learned representation in the tested model, rather than removal of an original variable from the forecasting dataset.

\subsection{Representation-Level Pre-Encoder Gate}

For a common interface, the representation exposed at the pre-encoder boundary is written as
\begin{equation}
E_g \in \mathbb{R}^{B \times N_c \times D},
\end{equation}
where $N_c$ denotes the number of representation units and $D$ is the model dimension. The gate computes a scalar score for every representation unit. With $D_h=\max(\lfloor D/2\rfloor,1)$, which reduces to $D/2$ for the reported model dimensions, the gate is
\begin{equation}
\begin{aligned}
H_g &= \phi(E_g W_1+b_1),\\
S &= \sigma(H_g W_2+b_2),\\
\widetilde{E}_g &= S\odot E_g,
\end{aligned}
\label{eq:gate}
\end{equation}
where $H_g$ is the hidden gate activation, $S\in\mathbb{R}^{B\times N_c\times1}$ is the sigmoid admission-score tensor, $W_1\in\mathbb{R}^{D\times D_h}$ and $b_1\in\mathbb{R}^{D_h}$ are the first-layer parameters, and $W_2\in\mathbb{R}^{D_h\times1}$ and $b_2\in\mathbb{R}$ are the second-layer parameters. Here, $\phi(\cdot)$ is the Gaussian error linear unit (GELU) \cite{hendrycks2016gaussian}, $\sigma(\cdot)$ is the sigmoid function, and $\odot$ denotes element-wise multiplication. The biases are broadcast over the batch and representation dimensions, and $S$ is broadcast over the feature dimension in the last expression. The multiplication by $W_1$ and $W_2$ is along the last feature dimension, so the intermediate tensors have shapes $B\times N_c\times D_h$ and $B\times N_c\times1$, respectively. The reweighted covariate representation is then passed to the original encoder in the layout required by the corresponding backbone.

The gate is intentionally multiplicative and does not include a residual bypass inside the module. The learned score is therefore the explicit scaling factor applied to each representation unit before encoder processing. The backbone can still form residual representations in its own layers, but the input interface remains explicit.

\subsection{Usage-Regularized Admission Objective}

The same interface can be trained with a covariate-usage penalty. The forecasting loss is the standard MSE loss,
\begin{equation}
\loss_{\mathrm{MSE}} = \frac{1}{BH}\sum_{b=1}^{B}\sum_{t=1}^{H}
\left(\hat{y}_{b,t}-y_{b,t}\right)^2.
\end{equation}
Here, $y_{b,t}$ and $\hat{y}_{b,t}$ are the target and predicted values at forecast step $t$ for batch item $b$, and $B$ and $H$ are the batch size and prediction horizon defined above. Let $u_j$ denote the assigned usage weight of the $j$th representation unit. In the controlled admission analysis, we use a uniform usage setting, $u_j=1$, to measure whether the gate can reduce average covariate admission. The usage-regularized training objective is
\begin{equation}
\loss_{\mathrm{use}} =
\loss_{\mathrm{MSE}} + \lambda \frac{1}{B N_c}\sum_{b=1}^{B}\sum_{j=1}^{N_c} S_{b,j} u_j,
\label{eq:usage_gate}
\end{equation}
where $\lambda$ controls the strength of the usage penalty and $S_{b,j}$ is the admission score for the $j$th representation unit. When $\lambda=0$, the objective reduces to the standard forecasting loss used by the ungated and unpenalized gated variants. The reported usage-regularized experiments use this term to regulate average admission.

\subsection{Soft-Start Initialization}

A naive sigmoid gate can make early training unstable if its initial values are too close to zero or too close to one. A gate initialized near zero removes covariate information before the model has learned its relevance, while a gate initialized near one approaches full covariate admission and weakens the purpose of input-side regulation.

We therefore use a soft-start initialization that leaves the covariate path partially open at the beginning of training. The same initialization rule is used for all gated backbones unless otherwise stated. This setting keeps gradients available through the covariate representation while avoiding full initial reliance on all inputs. During training, the model can increase or decrease the scores assigned to representation units.

\subsection{Backbone Integration}

The same representation-level pre-encoder gate is used for all evaluated Transformer-based backbones. It can be attached as a plug-in interface to a Transformer-based backbone when the backbone exposes its covariate representation before encoder processing: the gate computes a sigmoid score used as an admission weight and reweights the representation. The reweighted representation then enters the original encoder and follows the baseline forecasting pipeline. The encoder and prediction head remain unchanged, and no backbone-specific scoring rule is introduced.

\begin{figure}[!t]
\centering
\includegraphics[width=\textwidth]{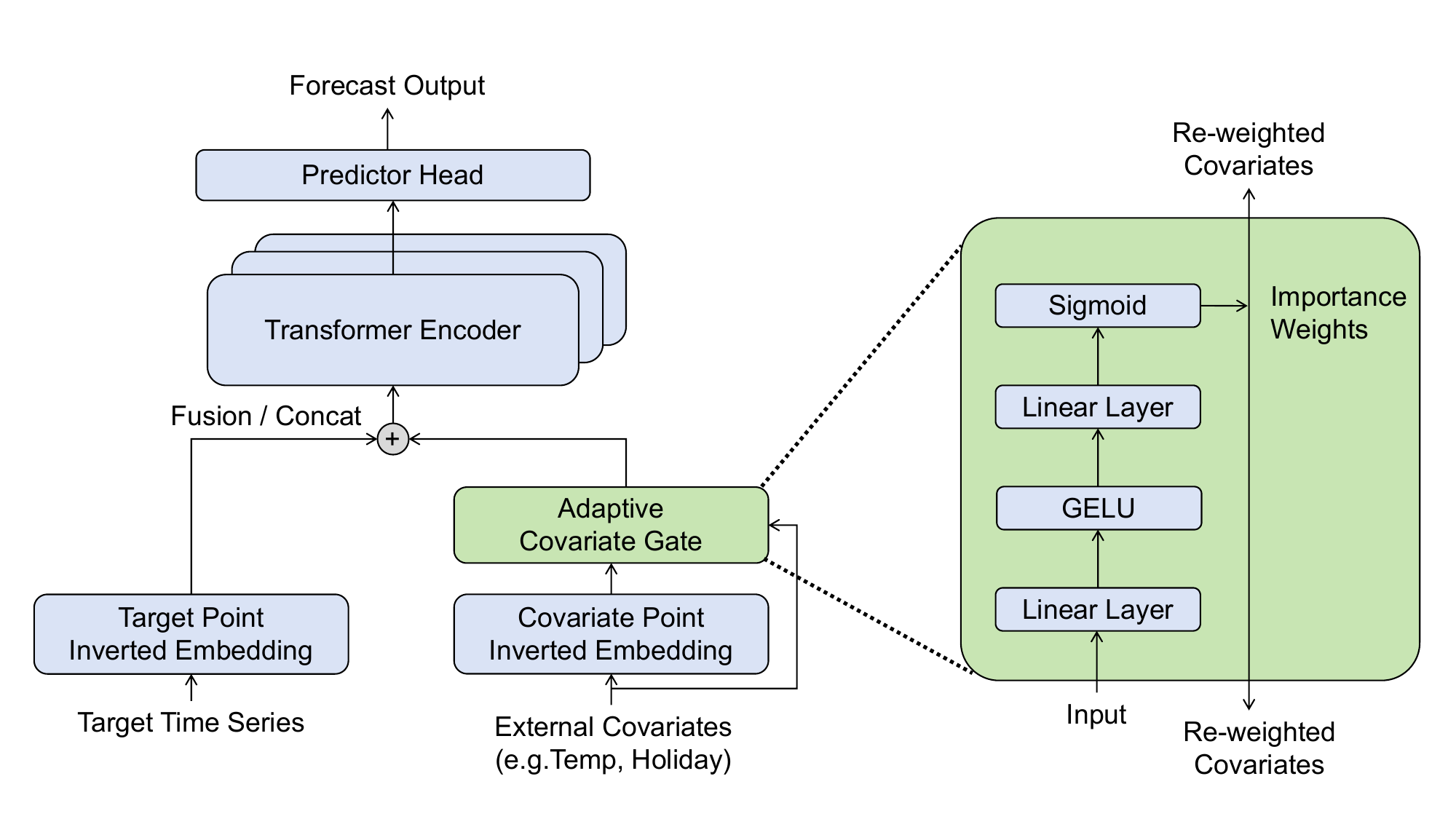}
\caption{Architecture of the pre-encoder covariate admission gate}
\label{fig:architecture}
\end{figure}

Fig.~\ref{fig:architecture} summarizes the general integration point. The design deliberately avoids backbone-specific scoring rules. The added computation consists of two linear layers and an element-wise multiplication over representation units. This keeps the tested change focused on the pre-encoder covariate interface.

\section{Experimental Setup}

\subsection{Task and Datasets}

We evaluate multivariate long-term forecasting. In each dataset, the last variable is treated as the target variable and the remaining variables are treated as covariates. The input construction, preprocessing, and forecasting task remain the same as in the corresponding baseline. The gate is inserted at the pre-encoder boundary of the covariate representation, following the exogenous-variable forecasting protocol used by TimeXer while retaining the common long-horizon benchmark setting \cite{wang2024timexer}.

Five datasets are used: \textit{Electricity Transformer Temperature minute-level} (\textit{ETTm1} and \textit{ETTm2}) datasets, \textit{Traffic}, \textit{Energy}, and \textit{influenza-like illness} (\textit{ILI}). These datasets follow commonly used long-term forecasting benchmarks \cite{lai2018modeling,zhou2021informer,wang2024timexer}. The \textit{ILI} dataset corresponds to the national influenza-like illness records in \texttt{national\_illness.csv}, which are derived from U.S. influenza surveillance reports provided through FluView. These datasets cover different covariate structures, including physical sensor variables, traffic measurements, energy-related variables, and public-health indicators. For \textit{ETTm1}, \textit{ETTm2}, \textit{Traffic}, and \textit{Energy}, the prediction horizons are 96, 192, 336, and 720. For \textit{ILI}, the horizons are 24, 36, 48, and 60. Mean squared error (MSE) and mean absolute error (MAE) are reported. MSE is more sensitive to larger forecast errors, whereas MAE reports the average absolute deviation, so the two metrics provide complementary views of forecasting behavior.

\subsection{Baselines and Training Protocol}

The evaluated backbones are TimeXer, iTransformer, and PatchTST. Each gated model inherits the corresponding baseline hyperparameters, including look-back length, model dimension, optimizer, learning rate, and batch size. The target variable, covariate set, preprocessing, and train-validation-test split are kept identical within each baseline--gate pair. No additional search is performed for the gate. The manuscript reports two implementation branches: the main implementation version for the paired baseline--gate comparison, and an auxiliary implementation version for the ablations, diagnostics, and contextual comparison.

The comparison is organized around baseline--gate pairs, and the reported values characterize the inserted interface under the existing backbone setting. This keeps the paper focused on whether a small representation-level pre-encoder gate can be attached to Transformer-based forecasting backbones without redesigning their training pipelines.

For TimeXer, we additionally include a placement-and-learnability ablation to isolate the design choice studied in this paper. Four TimeXer variants are compared: NoGate, which is the ungated TimeXer baseline; PreEncoderGate, which applies the learnable gate at the pre-encoder boundary; PostEncoderGate, which applies a gate after encoder processing; and FrozenGate, which uses the pre-encoder placement but freezes the gate parameters after initialization. This ablation is used to test whether the observed behavior is tied to the proposed pre-encoder location and to the learnability of the admission scores, rather than to adding an arbitrary multiplicative scaling layer.

MSE and MAE are reported throughout.

We also conduct an additional TimeXer analysis on \textit{ETTm1}, \textit{ETTm2}, and \textit{ILI}. This analysis compares how physical covariates enter the TimeXer covariate branch under four regimes. FullCov admits all physical covariates, TargetOnly removes the physical covariates and keeps the target forecasting setting without them, RandomK admits a fixed random subset of $k$ covariates, and VIFLowK admits the $k$ covariates with the lowest VIF values. RandomK samples without replacement using a dataset-stable random seed, and VIFLowK computes VIF values from the training portion of the physical covariates before sorting them in ascending order. UsageGate uses the usage-regularized objective in Eq.~\ref{eq:usage_gate}. The evaluated grid is $k\in\{1,2,3,4,5\}$ for the fixed-subset variants and $\lambda\in\{0,10^{-3},10^{-2},10^{-1}\}$ for UsageGate, with a uniform usage weight. Table~\ref{tab:controlled_admission} summarizes the comparison, and the appendix reports the full grids. This analysis compares the admission regimes under the same backbone and records the resulting tradeoff between average admission and forecasting error.

\section{Results and Analysis}

The results are organized around the covariate-admission interface. We first evaluate paired forecasting performance, then isolate placement and learnability, examine initialization sensitivity, test controlled covariate admission, and finally provide a redundancy diagnostic. A contextual comparison with forecasting baselines is reported after the main interface analyses.

\subsection{Paired Forecasting Performance}

Table~\ref{tab:main_results} reports paired gate--baseline comparisons under the main implementation version, averaged over fixed seeds. Each gated model inherits the corresponding baseline configuration, and no additional search is used for the gated branch. Bold values mark the lower mean within each gate--baseline pair, so the table reports paired results within each backbone rather than a global ranking across all methods.

Across the paired model--dataset--horizon settings, the gated variants are often at parity with, and in several cases better than, their paired baselines. On \textit{ETTm1} and \textit{ETTm2}, the gated variants mostly track their baselines closely. On \textit{Traffic}, TimeXer-Gate and PatchTST-Gate show the clearest gains, while iTransformer-Gate is mixed at longer horizons. On \textit{Energy}, the paired comparisons are near parity. On \textit{ILI}, the gate reduces MSE more consistently than MAE.

\begin{table}[!htbp]
\renewcommand{\arraystretch}{1.08}
\caption{Long-term forecasting performance}
\label{tab:main_results}
\centering
\scriptsize
\setlength{\tabcolsep}{2pt}
\begin{tabular*}{\linewidth}{@{\extracolsep{\fill}}lllccccc@{}}
\toprule
\multirow{2}{*}{Dataset} & \multirow{2}{*}{Method} & \multirow{2}{*}{Metric}
& \multicolumn{5}{c}{Horizon} \\
\cmidrule(lr){4-8}
& & & $H_1$ & $H_2$ & $H_3$ & $H_4$ & AVG \\
\midrule
\multirow{12}{*}{\textit{ETTm1}}
& TimeXer-Gate & MSE & \best{0.028} & \best{0.043} & \best{0.056} & \best{0.079} & \best{0.052} \\
& & MAE & \best{0.125} & \best{0.158} & \best{0.183} & \best{0.217} & \best{0.171} \\
& TimeXer & MSE & 0.028 & 0.043 & 0.057 & 0.079 & 0.052 \\
& & MAE & 0.125 & 0.158 & 0.185 & 0.217 & 0.171 \\
& iTransformer-Gate & MSE & \best{0.028} & \best{0.044} & \best{0.058} & 0.080 & \best{0.053} \\
& & MAE & \best{0.126} & \best{0.160} & \best{0.186} & 0.217 & \best{0.172} \\
& iTransformer & MSE & 0.029 & 0.044 & 0.059 & \best{0.080} & 0.053 \\
& & MAE & 0.127 & 0.161 & 0.187 & \best{0.217} & 0.173 \\
& PatchTST-Gate & MSE & \best{0.029} & \best{0.043} & \best{0.057} & \best{0.081} & \best{0.053} \\
& & MAE & \best{0.126} & \best{0.158} & \best{0.184} & \best{0.218} & \best{0.171} \\
& PatchTST & MSE & 0.029 & 0.043 & 0.057 & 0.081 & 0.053 \\
& & MAE & 0.126 & 0.159 & 0.184 & 0.219 & 0.172 \\
\midrule
\multirow{12}{*}{\textit{ETTm2}}
& TimeXer-Gate & MSE & \best{0.066} & \best{0.099} & \best{0.130} & \best{0.182} & \best{0.119} \\
& & MAE & \best{0.185} & \best{0.233} & \best{0.274} & \best{0.331} & \best{0.256} \\
& TimeXer & MSE & 0.067 & 0.102 & 0.131 & 0.182 & 0.121 \\
& & MAE & 0.187 & 0.238 & 0.275 & 0.332 & 0.258 \\
& iTransformer-Gate & MSE & \best{0.068} & \best{0.103} & \best{0.131} & \best{0.184} & \best{0.121} \\
& & MAE & \best{0.190} & \best{0.240} & \best{0.276} & \best{0.334} & \best{0.260} \\
& iTransformer & MSE & 0.075 & 0.110 & 0.131 & 0.184 & 0.125 \\
& & MAE & 0.200 & 0.250 & 0.277 & 0.335 & 0.265 \\
& PatchTST-Gate & MSE & 0.065 & \best{0.099} & 0.130 & \best{0.182} & \best{0.119} \\
& & MAE & \best{0.182} & \best{0.233} & 0.274 & \best{0.331} & \best{0.255} \\
& PatchTST & MSE & \best{0.065} & 0.101 & \best{0.130} & 0.182 & 0.119 \\
& & MAE & 0.183 & 0.236 & \best{0.274} & 0.331 & 0.256 \\
\midrule
\multirow{12}{*}{\textit{Traffic}}
& TimeXer-Gate & MSE & \best{0.149} & \best{0.150} & \best{0.149} & \best{0.169} & \best{0.154} \\
& & MAE & 0.224 & \best{0.227} & 0.232 & \best{0.252} & \best{0.234} \\
& TimeXer & MSE & 0.150 & 0.152 & 0.150 & 0.170 & 0.156 \\
& & MAE & \best{0.224} & 0.228 & \best{0.232} & 0.253 & 0.234 \\
& iTransformer-Gate & MSE & \best{0.147} & \best{0.146} & 0.145 & 0.167 & \best{0.151} \\
& & MAE & \best{0.225} & \best{0.224} & 0.229 & 0.253 & 0.233 \\
& iTransformer & MSE & 0.148 & 0.146 & \best{0.144} & \best{0.166} & 0.151 \\
& & MAE & 0.226 & 0.224 & \best{0.228} & \best{0.252} & \best{0.232} \\
& PatchTST-Gate & MSE & \best{0.174} & \best{0.169} & \best{0.167} & \best{0.196} & \best{0.177} \\
& & MAE & \best{0.252} & \best{0.247} & \best{0.248} & \best{0.271} & \best{0.254} \\
& PatchTST & MSE & 0.210 & 0.191 & 0.211 & 0.376 & 0.247 \\
& & MAE & 0.272 & 0.259 & 0.268 & 0.319 & 0.280 \\
\midrule
\multirow{12}{*}{\textit{Energy}}
& TimeXer-Gate & MSE & \best{0.214} & 0.244 & \best{0.274} & 0.298 & \best{0.258} \\
& & MAE & \best{0.323} & 0.347 & \best{0.368} & 0.405 & \best{0.361} \\
& TimeXer & MSE & 0.218 & \best{0.243} & 0.275 & \best{0.298} & 0.258 \\
& & MAE & 0.327 & \best{0.345} & 0.370 & \best{0.403} & 0.361 \\
& iTransformer-Gate & MSE & \best{0.221} & \best{0.250} & \best{0.284} & \best{0.313} & \best{0.267} \\
& & MAE & \best{0.327} & \best{0.349} & 0.376 & \best{0.415} & \best{0.367} \\
& iTransformer & MSE & 0.224 & 0.252 & 0.285 & 0.322 & 0.271 \\
& & MAE & 0.330 & 0.353 & \best{0.376} & 0.421 & 0.370 \\
& PatchTST-Gate & MSE & 0.210 & 0.244 & 0.282 & 0.304 & 0.260 \\
& & MAE & 0.321 & \best{0.345} & 0.374 & 0.406 & 0.361 \\
& PatchTST & MSE & \best{0.204} & \best{0.242} & \best{0.276} & \best{0.298} & \best{0.255} \\
& & MAE & \best{0.317} & 0.345 & \best{0.372} & \best{0.403} & \best{0.359} \\
\bottomrule
\end{tabular*}
\end{table}

\clearpage
\addtocounter{table}{-1}
\begin{table}[!htbp]
\renewcommand{\arraystretch}{1.08}
\caption{Long-term forecasting performance (continued)}
\centering
\scriptsize
\setlength{\tabcolsep}{2pt}
\begin{tabular*}{\linewidth}{@{\extracolsep{\fill}}lllccccc@{}}
\toprule
\multirow{2}{*}{Dataset} & \multirow{2}{*}{Method} & \multirow{2}{*}{Metric}
& \multicolumn{5}{c}{Horizon} \\
\cmidrule(lr){4-8}
& & & $H_1$ & $H_2$ & $H_3$ & $H_4$ & AVG \\
\midrule
\multirow{12}{*}{\textit{ILI}}
& TimeXer-Gate & MSE & \best{0.674} & \best{0.689} & 0.718 & \best{0.748} & \best{0.707} \\
& & MAE & 0.613 & 0.654 & 0.688 & \best{0.716} & 0.668 \\
& TimeXer & MSE & 0.686 & 0.702 & \best{0.717} & 0.754 & 0.715 \\
& & MAE & \best{0.609} & \best{0.653} & \best{0.681} & 0.718 & \best{0.665} \\
& iTransformer-Gate & MSE & \best{0.673} & \best{0.710} & \best{0.700} & \best{0.729} & \best{0.703} \\
& & MAE & 0.598 & \best{0.653} & \best{0.660} & \best{0.698} & \best{0.652} \\
& iTransformer & MSE & 0.674 & 0.715 & 0.702 & 0.730 & 0.705 \\
& & MAE & \best{0.596} & 0.655 & 0.662 & 0.700 & 0.653 \\
& PatchTST-Gate & MSE & \best{0.678} & \best{0.679} & \best{0.704} & 0.738 & \best{0.700} \\
& & MAE & \best{0.599} & \best{0.633} & \best{0.668} & \best{0.707} & \best{0.652} \\
& PatchTST & MSE & 0.681 & 0.684 & 0.707 & \best{0.737} & 0.702 \\
& & MAE & 0.605 & 0.635 & 0.669 & 0.707 & 0.654 \\
\bottomrule
\end{tabular*}
\par\smallskip
\footnotesize Lower MSE and MAE indicate better forecasting performance. Values are averaged over the fixed seeds used for the main implementation version. Bold values mark the lower mean within each gate--baseline pair. The table reports paired baseline--gate behavior under inherited configurations. For \textit{ETTm1}, \textit{ETTm2}, \textit{Traffic}, and \textit{Energy}, $H_1$--$H_4$ denote horizons 96, 192, 336, and 720; for \textit{ILI}, they denote horizons 24, 36, 48, and 60.
\end{table}

The zero-extra-tuning setting keeps the tuning budget aligned between each baseline and its gated counterpart. The seed-averaged results are reported as paired observations for the tested interface.

\subsection{TimeXer Ablation on Gate Placement and Learnability}

The paired results above evaluate the proposed gate against ungated backbones. To further examine whether the TimeXer improvement comes from the proposed pre-encoder interface, we compare four TimeXer variants in Table~\ref{tab:timexer_ablation}. NoGate is the original TimeXer baseline. PreEncoderGate is the proposed learnable pre-encoder gate. PostEncoderGate applies a gate after encoder processing, and FrozenGate keeps the pre-encoder placement but freezes the gate parameters after initialization. Thus, PostEncoderGate tests the effect of placement, while FrozenGate tests whether fixed covariate attenuation can explain the observed behavior.

\begin{table}[!htbp]
\renewcommand{\arraystretch}{1.08}
\caption{TimeXer ablation on gate placement and learnability}
\label{tab:timexer_ablation}
\centering
\scriptsize
\setlength{\tabcolsep}{2pt}
\begin{tabular*}{\linewidth}{@{\extracolsep{\fill}}lcccccccc@{}}
\toprule
\multirow{2}{*}{Dataset}
& \multicolumn{2}{c}{NoGate}
& \multicolumn{2}{c}{PreEncoderGate}
& \multicolumn{2}{c}{PostEncoderGate}
& \multicolumn{2}{c}{FrozenGate} \\
\cmidrule(lr){2-3}\cmidrule(lr){4-5}\cmidrule(lr){6-7}\cmidrule(lr){8-9}
& MSE & MAE & MSE & MAE & MSE & MAE & MSE & MAE \\
\midrule
\textit{ETTm1}
& 0.052 & 0.171 & \best{0.052} & \best{0.171} & 0.052 & 0.172 & 0.052 & 0.171 \\
\textit{ETTm2}
& 0.120 & 0.258 & \best{0.119} & \best{0.255} & 0.120 & 0.258 & 0.119 & 0.256 \\
\textit{Traffic}
& 0.156 & 0.235 & \best{0.153} & \best{0.233} & 0.168 & 0.255 & 0.160 & 0.242 \\
\textit{ILI}
& 0.726 & \best{0.670} & 0.719 & 0.673 & 0.770 & 0.714 & \best{0.719} & 0.673 \\
\textit{Energy}
& 0.259 & \best{0.361} & \best{0.258} & 0.361 & 0.259 & 0.362 & 0.258 & 0.362 \\
\midrule
Overall
& 0.263 & 0.339 & \best{0.260} & \best{0.339} & 0.274 & 0.352 & 0.262 & 0.341 \\
\bottomrule
\end{tabular*}
\par\smallskip
\footnotesize Lower values indicate better forecasting performance. MSE and MAE are compared across all four variants.
\end{table}

Across the reported averages, PreEncoderGate has the lowest MSE and tied lowest MAE among the four variants. PreEncoderGate improves over NoGate, while PostEncoderGate does not reproduce the same pattern. This comparison separates placement from learnability and identifies the pre-encoder learnable variant as the best-performing option in this ablation.

\subsection{Ablation on Gate Initialization}

The previous ablation examines where the gate should be placed and whether the gate should remain learnable. We further examine the soft-start initialization used by the proposed pre-encoder gate. This ablation keeps the TimeXer backbone, training protocol, datasets, horizons, and hyperparameters unchanged, and varies only the initial sigmoid admission setting. The original soft-start setting is compared with more open initial settings. These variants test whether the gain comes from simply admitting more covariate information at the beginning of training.

\begin{table}[!htbp]
\renewcommand{\arraystretch}{1.08}
\caption{TimeXer ablation on gate initialization}
\label{tab:gate_initialization}
\centering
\scriptsize
\setlength{\tabcolsep}{1.5pt}
\begin{tabular*}{\linewidth}{@{\extracolsep{\fill}}lcccccccccc@{}}
\toprule
\multirow{2}{*}{Dataset}
& \multicolumn{2}{c}{NoGate}
& \multicolumn{2}{c}{PreEncoderGate}
& \multicolumn{2}{c}{$p_0=0.500$}
& \multicolumn{2}{c}{$p_0=0.700$}
& \multicolumn{2}{c}{$p_0=0.900$} \\
\cmidrule(lr){2-3}\cmidrule(lr){4-5}\cmidrule(lr){6-7}\cmidrule(lr){8-9}\cmidrule(lr){10-11}
& MSE & MAE & MSE & MAE & MSE & MAE & MSE & MAE & MSE & MAE \\
\midrule
\textit{ETTm1}
& 0.052 & 0.171 & \best{0.052} & \best{0.171} & 0.052 & 0.171 & 0.052 & 0.171 & 0.052 & 0.171 \\
\textit{ETTm2}
& 0.120 & 0.258 & \best{0.119} & \best{0.255} & 0.119 & 0.256 & 0.120 & 0.257 & 0.120 & 0.257 \\
\textit{Energy}
& 0.259 & \best{0.361} & 0.258 & 0.361 & 0.259 & 0.361 & \best{0.257} & 0.361 & 0.258 & 0.363 \\
\textit{ILI}
& 0.726 & \best{0.670} & \best{0.719} & 0.673 & 0.722 & 0.672 & 0.725 & 0.672 & 0.727 & 0.672 \\
\midrule
Overall
& 0.289 & 0.365 & \best{0.287} & \best{0.365} & 0.288 & 0.365 & 0.288 & 0.365 & 0.289 & 0.366 \\
\bottomrule
\end{tabular*}
\par\smallskip
\footnotesize Lower MSE and MAE indicate better forecasting performance. The table reports averages over the tested horizons for each dataset. The initialization ablation is conducted on the four datasets available in this experiment, excluding \textit{Traffic}. $p_0$ denotes the initial sigmoid admission probability of the covariate gate.
\end{table}

Among the tested initializations, the original soft-start setting has the lowest overall MSE and tied lowest overall MAE. The more open variants do not improve the overall averages, so the soft-start choice remains the preferred initialization in this ablation.

\subsection{Controlled Covariate Admission Analysis}

The preceding ablations examine where the gate is placed and how it is initialized. We next compare four admission regimes within TimeXer: admitting all covariates, removing covariates, selecting a fixed subset, or learning a controllable pre-encoder admission score. Table~\ref{tab:controlled_admission} reports horizon-averaged results for TimeXer on \textit{ETTm1}, \textit{ETTm2}, and \textit{ILI}. FullCov admits all physical covariates, TargetOnly removes the physical covariates and keeps the target forecasting setting without them, RandomK admits a fixed random subset of $k$ covariates, VIFLowK admits the $k$ covariates with the lowest VIF values, and UsageGate uses the usage-regularized objective in Eq.~\ref{eq:usage_gate}. The average-admission-score column is one for FullCov, zero for TargetOnly, $k/C$ for fixed subsets, and the average learned admission score for UsageGate.

\begin{table}[!htbp]
\renewcommand{\arraystretch}{1.08}
\caption{Controlled covariate admission analysis with TimeXer}
\label{tab:controlled_admission}
\centering
\scriptsize
\setlength{\tabcolsep}{2pt}
\begin{tabular*}{\linewidth}{@{\extracolsep{\fill}}llcccc@{}}
\toprule
Dataset & Variant & Admission setting & Average admission score & MSE & MAE \\
\midrule
\multirow{6}{*}{\textit{ETTm1}}
& FullCov & all covariates & 1.000 & \best{0.052} & \best{0.171} \\
& TargetOnly & no physical covariates & 0.000 & 0.053 & 0.172 \\
& RandomK & random subset ($k=5$) & 0.833 & 0.052 & 0.172 \\
& VIFLowK & low-VIF subset ($k=5$) & 0.833 & 0.052 & 0.172 \\
& UsageGate & $\lambda=0$ & 0.427 & 0.053 & 0.173 \\
& UsageGate & penalty ($\lambda=10^{-1}$) & 0.003 & 0.053 & 0.172 \\
\midrule
\multirow{6}{*}{\textit{ETTm2}}
& FullCov & all covariates & 1.000 & 0.120 & 0.257 \\
& TargetOnly & no physical covariates & 0.000 & 0.121 & 0.258 \\
& RandomK & random subset ($k=5$) & 0.833 & 0.120 & 0.258 \\
& VIFLowK & low-VIF subset ($k=5$) & 0.833 & 0.120 & 0.258 \\
& UsageGate & $\lambda=0$ & 0.284 & 0.122 & 0.259 \\
& UsageGate & penalty ($\lambda=10^{-1}$) & 0.008 & \best{0.119} & \best{0.256} \\
\midrule
\multirow{6}{*}{\textit{ILI}}
& FullCov & all covariates & 1.000 & 0.728 & \best{0.671} \\
& TargetOnly & no physical covariates & 0.000 & \best{0.716} & 0.673 \\
& RandomK & random subset ($k=1$) & 0.167 & 0.722 & 0.673 \\
& VIFLowK & low-VIF subset ($k=1$) & 0.167 & 0.717 & 0.673 \\
& UsageGate & $\lambda=0$ & 0.274 & 0.719 & 0.673 \\
& UsageGate & penalty ($\lambda=10^{-1}$) & 0.253 & 0.719 & 0.673 \\
\bottomrule
\end{tabular*}
\par\smallskip
\footnotesize Lower MSE and MAE indicate better forecasting performance. Values are averaged over the four prediction horizons of each dataset. The summary table reports the observed results for the evaluated settings; the appendix reports the full grids. Average admission score denotes full admission for FullCov, no covariate admission for TargetOnly, the selected subset fraction for RandomK and VIFLowK, and the average learned admission score for UsageGate.
\end{table}

The table shows that admission regime matters. On \textit{ETTm1} and \textit{ETTm2}, FullCov is better than TargetOnly, so physical covariates carry useful information in these datasets. Fixed-subset results stay close to FullCov only when most covariates remain admitted, and neither RandomK nor VIFLowK is consistently superior in this table.

UsageGate provides a controllability check for the admission score. As $\lambda$ increases, the average learned admission score decreases, and the forecasting errors stay close to the full-covariate values in \textit{ETTm1} and \textit{ETTm2}. On \textit{ETTm2}, the penalized UsageGate setting has the lowest average MSE and MAE among the listed settings.

The \textit{ILI} result gives a different pattern. TargetOnly has the lowest MSE, while FullCov has the lowest MAE. The UsageGate variants do not beat TargetOnly on this dataset. Overall, the table shows that the same admission control can preserve accuracy in some settings while reducing average admission.

\subsection{Diagnostic Case Study on Covariate Redundancy}

Aggregate forecasting metrics indicate whether the gated variant changes prediction error. To inspect how the gate allocates admission weights across covariates, we include a diagnostic case study on one \textit{ETTm2} setting. The goal is to report a setting-specific association between learned gate weights, linear redundancy, and perturbation sensitivity, while keeping the main forecasting evidence in Table~\ref{tab:main_results}.

To inspect covariate behavior beyond aggregate forecasting error, we analyze one \textit{ETTm2} setting with look-back length \texttt{seq\_len=96} and prediction horizon \texttt{pred\_len=192}. Oil temperature (OT) is the target variable, and the remaining physical variables are treated as covariates. In the ETT notation, these covariates are high useful load (HUFL), high useless load (HULL), middle useful load (MUFL), middle useless load (MULL), low useful load (LUFL), and low useless load (LULL). We compute VIF to describe linear redundancy within the covariate set and use it to organize the perturbation analysis. For the $j$th covariate $x_j$, let $R_j^2$ be the coefficient of determination obtained by regressing $x_j$ on all remaining covariates. The VIF is
\begin{equation}
\mathrm{VIF}_j = \frac{1}{1-R_j^2}.
\label{eq:vif}
\end{equation}
Larger VIF values indicate that the covariate is easier to reconstruct from the rest of the covariate set. In this study, VIF describes linear redundancy in the covariate space.

The VIF calculation identifies HULL and MULL as the higher-VIF covariates in this setting. We therefore include both single-covariate perturbations and a grouped perturbation for this pair in the PFI analysis. PFI is related to permutation importance in random forests and model-reliance analysis, and measures how much the prediction error changes when a covariate is perturbed \cite{breiman2001random,fisher2019all}. For covariate $j$, PFI is defined as
\begin{equation}
\mathrm{PFI}_j =
\loss(f(X_{\mathrm{perm}(j)}),Y) -
\loss(f(X),Y),
\label{eq:pfi}
\end{equation}
where $X_{\mathrm{perm}(j)}$ denotes the input after randomly permuting the $j$th covariate. A positive value indicates that perturbing the covariate increases forecasting error, while a small or negative value indicates low sensitivity under this perturbation protocol. Because correlated covariates can substitute for each other, we also evaluate a grouped PFI perturbation for the high-VIF pair HULL and MULL. The PFI table reports responses measured by MSE, MAE, and Dynamic Time Warping (DTW) \cite{sakoe1978dynamic}.

The diagnostic analysis provides an additional view of the trained gate in this representative setting, especially for covariates that are strongly related to other covariates in the same input window.

\begin{figure}[!t]
\centering
\includegraphics[width=\textwidth]{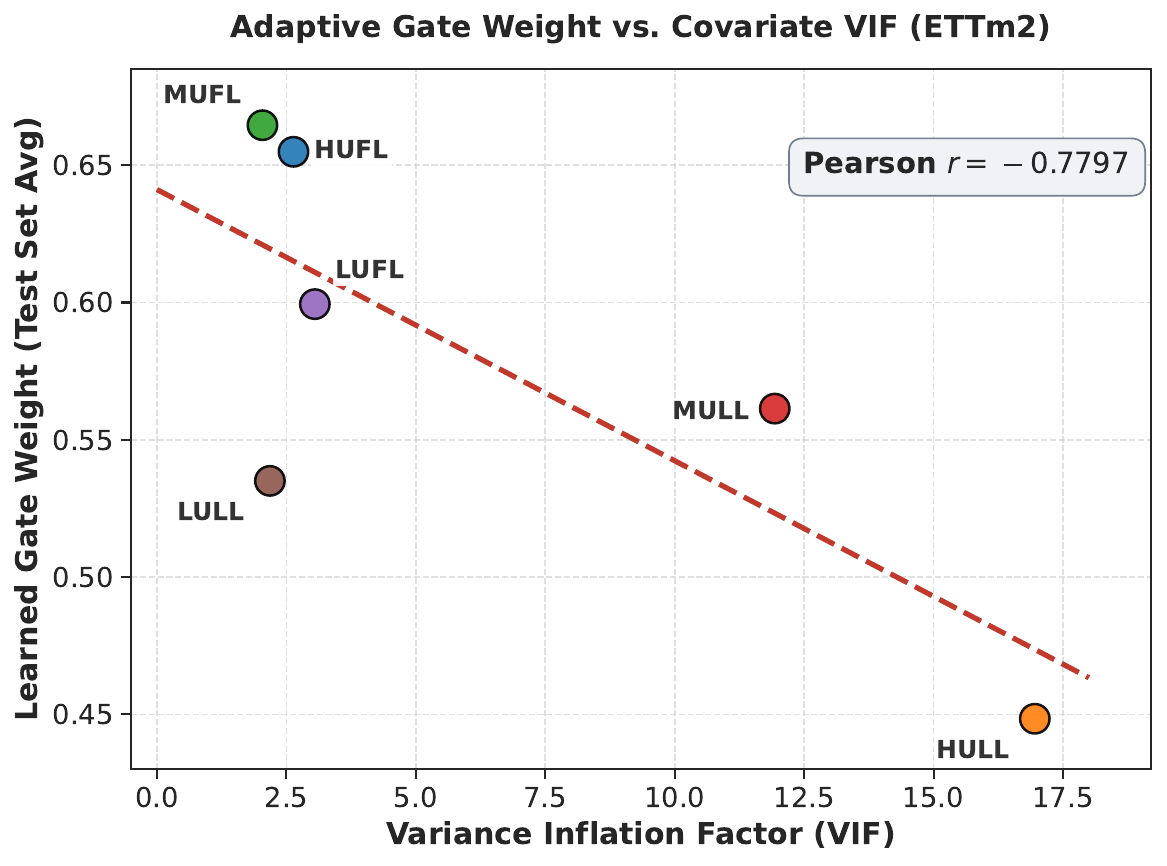}
\caption{Learned gate weight versus VIF on \textit{ETTm2}. The plot summarizes the relation between linear covariate redundancy and average gate weight in the analyzed setting.}
\label{fig:gate_vif}
\end{figure}

Table~\ref{tab:pfi} reports the PFI results together with the VIF values used to define the high-VIF group, and Fig.~\ref{fig:gate_vif} visualizes the setting-specific relation between VIF and learned gate weight. In this setting, the mean learned gate weight for the high-VIF group is lower than the mean for the remaining covariates. The PFI values provide a complementary view of model sensitivity under covariate perturbation. Together, these results provide a setting-specific diagnostic of how learned admission weights and perturbation responses vary across covariates.

\begin{table}[!htbp]
\renewcommand{\arraystretch}{1.08}
\caption{PFI perturbation analysis on \textit{ETTm2}}
\label{tab:pfi}
\centering
\scriptsize
\setlength{\tabcolsep}{2pt}
\begin{tabular*}{\linewidth}{@{\extracolsep{\fill}}lcclccc@{}}
\toprule
\multirow{2}{*}{Covariate/Group} & \multirow{2}{*}{VIF} & \multirow{2}{*}{Gate Weight}
& \multirow{2}{*}{Model} & \multicolumn{3}{c}{PFI} \\
\cmidrule(lr){5-7}
& & & & MSE & MAE & DTW \\
\midrule
\multirow{2}{*}{HUFL} & \multirow{2}{*}{2.637} & 0.655 & TimeXer-Gate & 0.000 & 0.000 & 0.012 \\
& & -- & TimeXer & 0.001 & 0.001 & 0.023 \\
\multirow{2}{*}{HULL} & \multirow{2}{*}{16.956} & 0.448 & TimeXer-Gate & 0.000 & 0.000 & 0.004 \\
& & -- & TimeXer & 0.000 & 0.000 & 0.004 \\
\multirow{2}{*}{MUFL} & \multirow{2}{*}{2.038} & 0.665 & TimeXer-Gate & 0.001 & 0.001 & 0.026 \\
& & -- & TimeXer & 0.001 & 0.001 & 0.027 \\
\multirow{2}{*}{MULL} & \multirow{2}{*}{11.933} & 0.561 & TimeXer-Gate & 0.000 & 0.000 & -0.008 \\
& & -- & TimeXer & 0.000 & 0.000 & -0.009 \\
\multirow{2}{*}{LUFL} & \multirow{2}{*}{3.050} & 0.599 & TimeXer-Gate & 0.001 & 0.002 & 0.020 \\
& & -- & TimeXer & 0.001 & 0.001 & 0.022 \\
\multirow{2}{*}{LULL} & \multirow{2}{*}{2.182} & 0.535 & TimeXer-Gate & 0.000 & 0.001 & 0.012 \\
& & -- & TimeXer & 0.001 & 0.001 & 0.019 \\
\midrule
\multirow{2}{*}{HULL+MULL} & \multirow{2}{*}{High-VIF} & -- & TimeXer-Gate & 0.000 & 0.000 & -0.007 \\
& & -- & TimeXer & 0.000 & 0.000 & -0.006 \\
\bottomrule
\end{tabular*}
\par\smallskip
\footnotesize The table reports PFI responses for TimeXer-Gate and TimeXer in the analyzed \textit{ETTm2} setting. The mean learned gate weight for the high-VIF group is lower than that for the remaining covariates in this case study. Gate Weight is reported for TimeXer-Gate; ``--'' marks the baseline rows without a learned gate. The grouped HULL+MULL perturbation reports the model response when the high-redundancy pair is perturbed together.
\end{table}

\subsection{Contextual Comparison with Forecasting Baselines}

Table~\ref{tab:recent_models} compares TimeXer-Gate with representative forecasting baselines from the result sheet, including TimeXer \cite{wang2024timexer}, iTransformer \cite{liu2024itransformer}, RLinear \cite{li2026revisiting}, PatchTST \cite{nie2022time}, TiDE \cite{das2023long}, TimesNet \cite{wu2022timesnet}, and DLinear \cite{zeng2023transformers}. Table~\ref{tab:recent_models} reports a contextual comparison under the auxiliary implementation version and a fixed seed: all methods use the same data split, preprocessing, and evaluation protocol, while each baseline retains its documented model-specific configuration. The comparison is restricted to \textit{ETTm1}, \textit{ETTm2}, and \textit{Traffic}, which overlap with the main experimental setting of this paper. This table contextualizes absolute forecasting performance, while the paired baseline--gate comparisons remain the main evidence for the admission interface.

On \textit{ETTm1}, TimeXer-Gate has the lowest average MSE and MAE in the table, with the largest differences appearing at the longer horizons. On \textit{ETTm2}, TimeXer-Gate also gives the lowest average result, although PatchTST is lower at horizon 336 and RLinear and TiDE are lower at horizon 720. On \textit{Traffic}, TimeXer-Gate gives the lowest average MSE and MAE in the table, while TiDE has the lowest MAE at horizon 336. These results provide context for the paired analyses above.

\begin{table}[!htbp]
\renewcommand{\arraystretch}{1.08}
\caption{TimeXer-Gate comparison with forecasting baselines}
\label{tab:recent_models}
\centering
\scriptsize
\setlength{\tabcolsep}{1.5pt}
\begin{tabular*}{\linewidth}{@{\extracolsep{\fill}}lllccccc@{}}
\toprule
\multirow{2}{*}{Dataset} & \multirow{2}{*}{Method} & \multirow{2}{*}{Metric}
& \multicolumn{5}{c}{Horizon} \\
\cmidrule(lr){4-8}
& & & $H_1$ & $H_2$ & $H_3$ & $H_4$ & AVG \\
\midrule
\multirow{16}{*}{\textit{ETTm1}}
& TimeXer-Gate & MSE & \best{0.028} & \best{0.043} & \best{0.056} & \best{0.079} & \best{0.052} \\
& & MAE & \best{0.125} & \best{0.158} & \best{0.183} & \best{0.216} & \best{0.171} \\
& TimeXer & MSE & \best{0.028} & \best{0.043} & 0.058 & \best{0.079} & 0.052 \\
& & MAE & \best{0.125} & \best{0.158} & 0.185 & 0.217 & 0.171 \\
& iTransformer & MSE & 0.029 & 0.045 & 0.060 & \best{0.079} & 0.053 \\
& & MAE & 0.128 & 0.163 & 0.190 & 0.218 & 0.175 \\
& RLinear & MSE & 0.030 & 0.044 & 0.057 & 0.080 & 0.053 \\
& & MAE & 0.129 & 0.160 & 0.184 & 0.217 & 0.173 \\
& PatchTST & MSE & 0.029 & 0.045 & 0.058 & 0.082 & 0.054 \\
& & MAE & 0.126 & 0.160 & 0.184 & 0.221 & 0.173 \\
& TiDE & MSE & 0.030 & 0.044 & 0.057 & 0.080 & 0.053 \\
& & MAE & 0.129 & 0.160 & 0.184 & 0.217 & 0.173 \\
& TimesNet & MSE & 0.029 & 0.044 & 0.061 & 0.083 & 0.054 \\
& & MAE & 0.128 & 0.160 & 0.190 & 0.223 & 0.175 \\
& DLinear & MSE & 0.034 & 0.055 & 0.078 & 0.098 & 0.066 \\
& & MAE & 0.135 & 0.173 & 0.210 & 0.234 & 0.188 \\
\midrule
\multirow{16}{*}{\textit{ETTm2}}
& TimeXer-Gate & MSE & \best{0.066} & \best{0.098} & 0.130 & 0.181 & \best{0.119} \\
& & MAE & \best{0.185} & \best{0.232} & 0.274 & 0.331 & \best{0.255} \\
& TimeXer & MSE & 0.067 & 0.101 & 0.130 & 0.182 & 0.120 \\
& & MAE & 0.188 & 0.236 & 0.275 & 0.332 & 0.258 \\
& iTransformer & MSE & 0.071 & 0.108 & 0.140 & 0.188 & 0.127 \\
& & MAE & 0.194 & 0.247 & 0.288 & 0.340 & 0.267 \\
& RLinear & MSE & 0.074 & 0.104 & 0.131 & \best{0.180} & 0.122 \\
& & MAE & 0.199 & 0.241 & 0.276 & \best{0.329} & 0.261 \\
& PatchTST & MSE & 0.068 & 0.100 & \best{0.128} & 0.185 & 0.120 \\
& & MAE & 0.188 & 0.236 & \best{0.271} & 0.335 & 0.258 \\
& TiDE & MSE & 0.073 & 0.104 & 0.131 & \best{0.180} & 0.122 \\
& & MAE & 0.199 & 0.241 & 0.276 & \best{0.329} & 0.261 \\
& TimesNet & MSE & 0.073 & 0.106 & 0.150 & 0.186 & 0.129 \\
& & MAE & 0.200 & 0.247 & 0.296 & 0.338 & 0.270 \\
& DLinear & MSE & 0.072 & 0.105 & 0.136 & 0.191 & 0.126 \\
& & MAE & 0.195 & 0.240 & 0.280 & 0.335 & 0.263 \\
\bottomrule
\end{tabular*}
\end{table}

\clearpage
\addtocounter{table}{-1}
\begin{table}[!htbp]
\renewcommand{\arraystretch}{1.08}
\caption{TimeXer-Gate comparison with forecasting baselines (continued)}
\centering
\scriptsize
\setlength{\tabcolsep}{1.5pt}
\begin{tabular*}{\linewidth}{@{\extracolsep{\fill}}lllccccc@{}}
\toprule
\multirow{2}{*}{Dataset} & \multirow{2}{*}{Method} & \multirow{2}{*}{Metric}
& \multicolumn{5}{c}{Horizon} \\
\cmidrule(lr){4-8}
& & & $H_1$ & $H_2$ & $H_3$ & $H_4$ & AVG \\
\midrule
\multirow{16}{*}{\textit{Traffic}}
& TimeXer-Gate & MSE & \best{0.149} & \best{0.150} & \best{0.148} & \best{0.165} & \best{0.153} \\
& & MAE & \best{0.224} & \best{0.229} & 0.230 & \best{0.247} & \best{0.232} \\
& TimeXer & MSE & 0.151 & 0.152 & 0.150 & 0.172 & 0.156 \\
& & MAE & \best{0.224} & \best{0.229} & 0.232 & 0.253 & 0.235 \\
& iTransformer & MSE & 0.156 & 0.156 & 0.154 & 0.177 & 0.161 \\
& & MAE & 0.236 & 0.237 & 0.243 & 0.268 & 0.246 \\
& RLinear & MSE & 0.350 & 0.314 & 0.305 & 0.328 & 0.324 \\
& & MAE & 0.431 & 0.404 & 0.399 & 0.415 & 0.412 \\
& PatchTST & MSE & 0.176 & 0.162 & 0.164 & 0.189 & 0.173 \\
& & MAE & 0.253 & 0.243 & 0.248 & 0.267 & 0.253 \\
& TiDE & MSE & 0.350 & 0.230 & 0.220 & 0.243 & 0.261 \\
& & MAE & 0.430 & 0.315 & \best{0.208} & 0.329 & 0.321 \\
& TimesNet & MSE & 0.154 & 0.164 & 0.167 & 0.197 & 0.171 \\
& & MAE & 0.249 & 0.255 & 0.259 & 0.292 & 0.264 \\
& DLinear & MSE & 0.268 & 0.302 & 0.298 & 0.340 & 0.302 \\
& & MAE & 0.351 & 0.387 & 0.384 & 0.416 & 0.385 \\
\bottomrule
\end{tabular*}
\par\smallskip
\footnotesize Lower MSE and MAE indicate better forecasting performance. Bold values mark the lowest entry for each metric in the same dataset and horizon. All methods use a fixed seed under the auxiliary implementation version and shared task configuration, data split, preprocessing, and evaluation protocol; each baseline retains its documented model-specific configuration. Values are transcribed from the result sheet and shown to three decimal places where needed. The table is a contextual comparison on overlapping datasets; the main paired comparison is reported in Table~\ref{tab:main_results}. For \textit{ETTm1}, \textit{ETTm2}, and \textit{Traffic}, $H_1$--$H_4$ denote horizons 96, 192, 336, and 720.
\end{table}

\subsection{Limitations}

The proposed gate is a lightweight representation-level pre-encoder module, and its effect depends on the dataset and the backbone. The results on \textit{Energy} and some evaluated backbone settings show that the same interface does not always reduce forecasting error.

The zero-extra-tuning protocol keeps the comparison aligned between each baseline and its gated counterpart, but it also constrains the gated variants to the inherited learning rate, model dimension, and training schedule.

The controlled usage penalty is a benchmark-level proxy, not a real acquisition-cost model. The VIF-informed PFI analysis is likewise a diagnostic view of one redundant-covariate setting.

\section{Conclusion}

This paper frames pre-encoder covariate admission as an interface problem in covariate-rich long-term forecasting. The proposed representation-level gate is placed after a baseline constructs its covariate representation and before the first encoder layer, so the covariate path can be regulated without changing the encoder or prediction head. The method adds a lightweight admission module and supports a controlled covariate-usage penalty.

Across the reported paired settings, the gate is competitive with the corresponding baselines and produces modest gains in several cases. The TimeXer ablations separate the effects of placement, learnability, and initialization. The controlled covariate-admission analysis shows that the learned admission score can be reduced under the usage penalty while keeping error close to the unpenalized setting in the tested TimeXer configurations. The VIF-informed PFI case study gives a setting-specific diagnostic of redundancy and perturbation sensitivity. Overall, the proposed gate is a compact plug-in interface for regulating covariate usage in Transformer-based backbones under the reported benchmarks.

\clearpage

\section*{Statements and Declarations}

\subsection*{Funding}

Anonymous for peer review.

\subsection*{Competing interests}

The authors have no competing interests to declare that are relevant to the content of this article.

\subsection*{Ethics approval}

Not applicable.

\subsection*{Data availability}

The ETTm1, ETTm2, Traffic, and influenza-like illness (ILI) datasets analysed in this study are publicly available benchmark datasets. The ETTm1, ETTm2, Traffic, and ILI benchmark files are publicly available through the Time-Series-Library data distribution (\url{https://github.com/thuml/Time-Series-Library}). The ILI benchmark data are derived from the publicly available U.S. Centers for Disease Control and Prevention FluView surveillance data (\url{https://www.cdc.gov/fluview/}). The Energy dataset, including \texttt{energy\_dataset.csv}, is publicly available in the ``Energy Consumption, Generation, Prices and Weather'' dataset on Kaggle (\url{https://www.kaggle.com/datasets/nicholasjhana/energy-consumption-generation-prices-and-weather/data?select=energy_dataset.csv}). No new datasets were generated during the current study.

\subsection*{Code availability}

The code supporting the findings of this study will be made publicly available upon publication.


\clearpage
\begin{appendices}
\section{Implementation Notes}

The gate adds only two linear layers to the representation exposed at the pre-encoder boundary. For the reported even model dimensions, the first layer contains $D(D/2)$ weights and $D/2$ biases, while the output layer contains $D/2$ weights and one bias. The additional parameter count is therefore $D^2/2 + D + 1$. This overhead is small relative to Transformer backbones. During inference, the gate requires one additional MLP pass over representation units and an element-wise multiplication.

\subsection{Controlled-Admission Grid Results}

The controlled-admission analysis evaluates the complete fixed-subset grid $k\in\{1,2,3,4,5\}$ and the formal UsageGate grid $\lambda\in\{0,10^{-3},10^{-2},10^{-1}\}$. Tables~\ref{tab:randomk_grid}--\ref{tab:usage_grid} report every evaluated setting used for the controlled comparison from the auxiliary implementation version. Each cell is written as MSE/MAE. For \textit{ETTm1} and \textit{ETTm2}, $H_1$--$H_4$ denote horizons 96, 192, 336, and 720; for \textit{ILI}, they denote horizons 24, 36, 48, and 60. AVG is the arithmetic mean over the four horizons. The rows shown in Table~\ref{tab:controlled_admission} are the settings with the lowest AVG MSE within the corresponding grid for each dataset.

\begin{table}[!htbp]
\renewcommand{\arraystretch}{1.05}
\caption{Random-subset grid results}
\label{tab:randomk_grid}
\centering
\scriptsize
\setlength{\tabcolsep}{2pt}
\begin{tabular*}{\linewidth}{@{\extracolsep{\fill}}llccccc@{}}
\toprule
Dataset & Setting & $H_1$ & $H_2$ & $H_3$ & $H_4$ & AVG \\
\midrule
\multirow{5}{*}{\textit{ETTm1}}
& $k=1$ & 0.029/0.127 & 0.045/0.161 & 0.059/0.186 & 0.081/0.218 & 0.053/0.173 \\
& $k=2$ & 0.029/0.126 & 0.044/0.161 & 0.058/0.186 & 0.080/0.217 & 0.053/0.173 \\
& $k=3$ & 0.028/0.125 & 0.044/0.160 & 0.058/0.186 & 0.080/0.217 & 0.052/0.172 \\
& $k=4$ & 0.028/0.125 & 0.044/0.159 & 0.058/0.185 & 0.080/0.217 & 0.052/0.172 \\
& $k=5$ & 0.028/0.125 & 0.043/0.158 & 0.058/0.185 & 0.079/0.217 & 0.052/0.172 \\
\midrule
\multirow{5}{*}{\textit{ETTm2}}
& $k=1$ & 0.069/0.189 & 0.107/0.242 & 0.135/0.278 & 0.183/0.333 & 0.123/0.260 \\
& $k=2$ & 0.070/0.190 & 0.109/0.244 & 0.135/0.279 & 0.184/0.333 & 0.124/0.261 \\
& $k=3$ & 0.068/0.188 & 0.106/0.241 & 0.132/0.277 & 0.182/0.331 & 0.122/0.259 \\
& $k=4$ & 0.067/0.186 & 0.104/0.238 & 0.132/0.276 & 0.182/0.331 & 0.121/0.258 \\
& $k=5$ & 0.067/0.187 & 0.103/0.237 & 0.131/0.275 & 0.181/0.331 & 0.120/0.258 \\
\midrule
\multirow{5}{*}{\textit{ILI}}
& $k=1$ & 0.708/0.626 & 0.706/0.659 & 0.722/0.689 & 0.754/0.720 & 0.722/0.673 \\
& $k=2$ & 0.715/0.625 & 0.713/0.659 & 0.724/0.688 & 0.757/0.721 & 0.727/0.673 \\
& $k=3$ & 0.717/0.625 & 0.717/0.659 & 0.724/0.687 & 0.759/0.722 & 0.729/0.673 \\
& $k=4$ & 0.717/0.623 & 0.718/0.659 & 0.725/0.686 & 0.759/0.722 & 0.730/0.673 \\
& $k=5$ & 0.718/0.623 & 0.720/0.659 & 0.725/0.685 & 0.760/0.722 & 0.731/0.672 \\
\bottomrule
\end{tabular*}
\par\smallskip
\footnotesize Each cell reports MSE/MAE; values are rounded to three decimal places from the final log metrics. AVG is computed over the four prediction horizons of the corresponding dataset.
\end{table}

\begin{table}[!htbp]
\renewcommand{\arraystretch}{1.05}
\caption{Low-VIF subset grid results}
\label{tab:viflowk_grid}
\centering
\scriptsize
\setlength{\tabcolsep}{2pt}
\begin{tabular*}{\linewidth}{@{\extracolsep{\fill}}llccccc@{}}
\toprule
Dataset & Setting & $H_1$ & $H_2$ & $H_3$ & $H_4$ & AVG \\
\midrule
\multirow{5}{*}{\textit{ETTm1}}
& $k=1$ & 0.029/0.127 & 0.045/0.161 & 0.059/0.186 & 0.081/0.218 & 0.053/0.173 \\
& $k=2$ & 0.029/0.127 & 0.044/0.160 & 0.059/0.187 & 0.081/0.218 & 0.053/0.173 \\
& $k=3$ & 0.029/0.127 & 0.044/0.160 & 0.058/0.187 & 0.080/0.218 & 0.053/0.173 \\
& $k=4$ & 0.029/0.127 & 0.044/0.160 & 0.058/0.186 & 0.080/0.218 & 0.053/0.173 \\
& $k=5$ & 0.028/0.126 & 0.044/0.159 & 0.058/0.186 & 0.079/0.217 & 0.052/0.172 \\
\midrule
\multirow{5}{*}{\textit{ETTm2}}
& $k=1$ & 0.067/0.186 & 0.105/0.239 & 0.134/0.278 & 0.182/0.331 & 0.122/0.259 \\
& $k=2$ & 0.069/0.187 & 0.105/0.239 & 0.134/0.278 & 0.183/0.332 & 0.123/0.259 \\
& $k=3$ & 0.068/0.189 & 0.102/0.236 & 0.131/0.275 & 0.181/0.331 & 0.121/0.258 \\
& $k=4$ & 0.067/0.187 & 0.103/0.238 & 0.131/0.276 & 0.181/0.331 & 0.121/0.258 \\
& $k=5$ & 0.068/0.189 & 0.101/0.235 & 0.131/0.276 & 0.181/0.331 & 0.120/0.258 \\
\midrule
\multirow{5}{*}{\textit{ILI}}
& $k=1$ & 0.699/0.625 & 0.700/0.658 & 0.719/0.689 & 0.751/0.719 & 0.717/0.673 \\
& $k=2$ & 0.705/0.623 & 0.705/0.657 & 0.721/0.687 & 0.754/0.719 & 0.721/0.672 \\
& $k=3$ & 0.709/0.622 & 0.710/0.657 & 0.721/0.686 & 0.756/0.720 & 0.724/0.671 \\
& $k=4$ & 0.711/0.622 & 0.713/0.658 & 0.723/0.685 & 0.759/0.721 & 0.726/0.671 \\
& $k=5$ & 0.712/0.622 & 0.715/0.658 & 0.723/0.685 & 0.760/0.722 & 0.728/0.672 \\
\bottomrule
\end{tabular*}
\par\smallskip
\footnotesize Each cell reports MSE/MAE; values are rounded to three decimal places from the final log metrics. AVG is computed over the four prediction horizons of the corresponding dataset.
\end{table}

\begin{table}[!htbp]
\renewcommand{\arraystretch}{1.05}
\caption{UsageGate penalty grid results}
\label{tab:usage_grid}
\centering
\scriptsize
\setlength{\tabcolsep}{2pt}
\begin{tabular*}{\linewidth}{@{\extracolsep{\fill}}llccccc@{}}
\toprule
Dataset & Setting & $H_1$ & $H_2$ & $H_3$ & $H_4$ & AVG \\
\midrule
\multirow{4}{*}{\textit{ETTm1}}
& $\lambda=0$ & 0.029/0.126 & 0.045/0.161 & 0.058/0.187 & 0.080/0.218 & 0.053/0.173 \\
& $\lambda=10^{-3}$ & 0.028/0.126 & 0.045/0.161 & 0.060/0.189 & 0.080/0.218 & 0.053/0.173 \\
& $\lambda=10^{-2}$ & 0.028/0.125 & 0.044/0.160 & 0.059/0.186 & 0.080/0.217 & 0.053/0.172 \\
& $\lambda=10^{-1}$ & 0.028/0.125 & 0.043/0.159 & 0.058/0.186 & 0.080/0.217 & 0.053/0.172 \\
\midrule
\multirow{4}{*}{\textit{ETTm2}}
& $\lambda=0$ & 0.066/0.185 & 0.104/0.239 & 0.134/0.278 & 0.183/0.333 & 0.122/0.259 \\
& $\lambda=10^{-3}$ & 0.067/0.186 & 0.104/0.238 & 0.133/0.276 & 0.181/0.331 & 0.121/0.258 \\
& $\lambda=10^{-2}$ & 0.065/0.184 & 0.103/0.238 & 0.132/0.275 & 0.180/0.329 & 0.120/0.257 \\
& $\lambda=10^{-1}$ & 0.065/0.183 & 0.104/0.239 & 0.131/0.275 & 0.178/0.327 & 0.119/0.256 \\
\midrule
\multirow{4}{*}{\textit{ILI}}
& $\lambda=0$ & 0.701/0.625 & 0.701/0.658 & 0.721/0.690 & 0.753/0.720 & 0.719/0.673 \\
& $\lambda=10^{-3}$ & 0.701/0.625 & 0.701/0.658 & 0.721/0.690 & 0.753/0.720 & 0.719/0.673 \\
& $\lambda=10^{-2}$ & 0.701/0.625 & 0.701/0.658 & 0.721/0.690 & 0.753/0.720 & 0.719/0.673 \\
& $\lambda=10^{-1}$ & 0.700/0.625 & 0.701/0.658 & 0.721/0.690 & 0.753/0.720 & 0.719/0.673 \\
\bottomrule
\end{tabular*}
\par\smallskip
\footnotesize Each cell reports MSE/MAE; values are rounded to three decimal places from the final log metrics. AVG is computed over the four prediction horizons of the corresponding dataset. The formal reported grid contains the four displayed $\lambda$ values.
\end{table}

\end{appendices}


\clearpage
\bibliography{sn-bibliography}

\end{document}